\documentclass[lettersize,journal]{IEEEtran}
\usepackage{amsmath,amsfonts}
\usepackage{algorithmic}
\usepackage{algorithm}
\usepackage{array}
\usepackage[caption=false,font=normalsize,labelfont=sf,textfont=sf]{subfig}
\usepackage{textcomp}
\usepackage{stfloats}
\usepackage{url}
\usepackage{verbatim}
\usepackage{graphicx}
\usepackage{cite}
\usepackage{siunitx}
\usepackage{amsmath,amssymb,amsfonts}
\usepackage{algorithmic}
\usepackage{wrapfig}
\usepackage{booktabs}
\usepackage{threeparttable}

\begin{document}

\title{Weakly Supervised Seafloor Segmentation for Seagrass Habitat Mapping in Side-Scan Sonar Imagery}

\author{Hayat Rajani,~\IEEEmembership{Member,~IEEE,}
        Nuno Gracias,~\IEEEmembership{Member,~IEEE,}
        and Rafael Garcia,~\IEEEmembership{Member,~IEEE}

\thanks{All authors are with the Computer Vision and Robotics Research Institute (ViCOROB) of the University of Girona, Spain.}

\thanks{Corresponding author: Hayat Rajani (hayat.rajani@udg.edu).}}



\maketitle

\begin{abstract}
Seagrass meadows are crucial blue-carbon habitats, and mapping their extent is a prerequisite for coastal management and carbon inventory. Optical satellite sensors cover large areas but cannot reach deep or turbid water, whereas side-scan sonar (SSS) images the seabed at high resolution and at any depth. Interpreting SSS, however, still relies on dense manual annotation, which is slow and costly. We address this by adapting a weakly supervised semantic segmentation framework to SSS benthic habitat mapping, so that pixel-level maps are learned from image-level labels alone. The framework couples a ViT-based encoder-decoder with a classification branch, extracts class activation maps, and refines them into pseudo-labels with a dense conditional random field that we tune for the noise and weak boundaries of acoustic imagery. It follows an iterative self-training scheme, together with a sampling strategy to cope with the strong class imbalance of the data. We also study the effect of different loss functions on segmentation quality, finding Lov\'asz-Softmax loss the most effective. On a held-out transect, the refined pseudo-labels reached an mIoU of 89.3\% against the ground truth, and the segmentation branch, trained without any pixel-level labels, reached 87.6\%. Self-supervised pretraining on unlabelled SSS added a further 3\% in mean intersection-over-union. Field trials further demonstrate the generalizability of the trained model. These results show that accurate and label-efficient benthic habitat mapping from side-scan sonar is feasible at the scale needed for coast-wide seagrass monitoring.
\end{abstract}

\begin{IEEEkeywords}
side-scan sonar, weakly-supervised semantic segmentation, seafloor segmentation, benthic habitat mapping, seagrass mapping, blue carbon footprint
\end{IEEEkeywords}


\section{Introduction}

\IEEEPARstart{S}{eagrass} meadows are among the most valuable coastal ecosystems. They provide habitat and nursery grounds, stabilise sediments, and store large amounts of organic carbon in their soils over long time scales, a service commonly termed blue carbon \cite{fourqurean2012seagrass}. Their global extent has declined by about 29\% since records began in 1879, driven by coastal development, declining water quality, and climate change \cite{waycott2009accelerating}. In the Mediterranean, the endemic seagrass \textit{Posidonia oceanica} forms extensive meadows that hold some of the largest sedimentary carbon stores in the basin, while \textit{Cymodocea nodosa} frequently occupies shallower, sandy, or disturbed areas alongside it \cite{fourqurean2012seagrass}. Reliable mapping of the spatial extent of these meadows is therefore a prerequisite for coastal management and for any inventory of the carbon they hold \cite{pasqualini1998mapping}. 



Mapping these meadows over large areas requires imaging the seabed at fine spatial resolution. Side-scan sonar (SSS) produces high-resolution acoustic imagery of the seabed and is widely used in underwater applications such as marine archaeology, structural inspection, and environmental monitoring \cite{blondel2009handbook}. It has long been applied to map Mediterranean seagrass meadows, including \textit{Posidonia oceanica}, at depths where optical sensors cannot operate \cite{pasqualini1998mapping}. Optical satellite and aerial sensors map shallow meadows well but cannot reach the deeper limits of these habitats, so acoustic mapping with SSS complements space-borne and airborne Earth observation rather than competing with it \cite{pasqualini1998mapping}. Unlike optical imagery, however, SSS lacks mature automated or semi-automated annotation tools. General-purpose segmenters such as the Segment Anything Model (SAM) \cite{kirillov2023segment} and curated underwater image databases such as FathomNet \cite{katija2022fathomnet} are built for optical data and do not transfer to acoustic imagery. As a result, interpreting SSS maps still relies on the manual annotation of terrain and structural features, which is labour-intensive and costly \cite{rajani2023}. Automating seabed classification would make large-area habitat mapping practical, and running it on board autonomous underwater vehicles (AUVs) would allow classification to proceed during the survey and support more informed, real-time decisions.


This work builds on our earlier work \cite{rajani2023}, in which we developed an encoder-decoder architecture that combines Vision Transformers (ViTs) and convolutional neural networks (CNNs) for semantic segmentation of the seafloor in SSS imagery. The model improved on the previous state-of-the-art by a large margin while meeting real-time computational requirements. Despite its strong performance and its generalisation to unseen data, it was trained under noisy supervision. The seafloor categorisations provided by geophysicists are often produced at a coarse level on SSS mosaics, whereas time-critical online processing must operate directly on raw SSS waterfalls. Transferring annotations from mosaics to waterfalls is not straightforward, especially without access to the internal parameters used for mosaicing, and this introduces discrepancies. The dense ground truth used for supervision is therefore not pixel-accurate. This setting can be treated as a form of weak supervision, in which training proceeds under noisy or incomplete labels.

Weakly supervised semantic segmentation (WSSS) aims to produce a pixel-level segmentation while relying only on coarse or inexpensive labels, for example image-level tags that state which classes appear in an image without indicating where they are \cite{chen2025weakly, ahn2018learning}. This lowers the annotation effort substantially, at the cost of a harder learning problem, and it is well suited to domains such as SSS where dense labels are scarce. This motivates the main objective of the present work, a segmentation pipeline for side-scan sonar aimed at label-efficient benthic habitat mapping at the scale needed for coast-wide seagrass monitoring.

In this work, we do not propose a new model. Instead, we adapt an existing WSSS framework and analyse the effect of label refinement and loss function design. Specifically, we tune the parameters of a dense conditional random field (CRF) \cite{krahenbuhl2011efficient} and compare segmentation performance under several loss functions, including cross-entropy, focal loss \cite{lin2017focal}, and Lov\'asz-Softmax loss \cite{berman2018lovasz}. Our analysis also shows that commonly used evaluation metrics such as the mean intersection-over-union (mIoU) can obscure meaningful differences between models, particularly under class imbalance. The improvements we report are better understood by interpreting the metric together with the qualitative segmentation behaviour.

We frame the study around three questions:
\begin{itemize}
\item Can a weakly supervised approach based on image-level labels produce accurate benthic habitat maps from SSS imagery?
\item How does it compare with fully supervised methods in terms of accuracy and efficiency?
\item What are its limitations for large-area seagrass and seafloor mapping, and how can they be addressed?
\item How well does the approach transfer across survey sites and acoustic conditions, as required for large-area habitat monitoring?
\end{itemize}

 
\section{Related Work}

Optical satellite and aerial sensors are the most common choice for seagrass mapping because they cover large areas at low cost. Traganos \textit{et al.} mapped seagrass extent across the entire Mediterranean from Sentinel-2 imagery and linked the result to blue carbon accounting \cite{traganos2022spatially}. Giménez-Romero \textit{et al.} trained a convolutional network to map \textit{Posidonia oceanica} meadows across the Mediterranean, with a focus on generalisation to new regions \cite{gimenez2025generalizable}. Jeon \textit{et al.} compared deep networks for the semantic segmentation of seagrass habitat from drone imagery \cite{jeon2021semantic}. These studies confirm that optical data support coast-wide monitoring, but they share a physical limit. Light is attenuated in the water column, so optical sensors map shallow meadows well but cannot reliably reach the deeper limit of \textit{Posidonia oceanica}, which extends to about 40\,\unit{m}. Additionally, the maps created by such optical systems tend to have a coarse spatial resolution, as in the 10\,\unit{m} pan-Mediterranean map of \cite{traganos2022spatially}, which is too coarse to resolve fine meadow boundaries or to separate sparse from dense cover.

Acoustic sensors fill this gap, because sound propagates well in turbid and deep water. Side-scan sonar (SSS) produces high-resolution backscatter imagery of the seabed at decimetre resolution and reaches depths that optical sensors cannot, thereby resolving structure that a 10\,\unit{m} pixel cannot. It has long been used for seagrass cartography. Pasqualini \textit{et al.} combined aerial photographs for shallow water with SSS for the deeper range to map \textit{Posidonia oceanica} off Corsica \cite{pasqualini1998mapping}, and later used SSS to support the management of Mediterranean littoral ecosystems \cite{pasqualini2000contribution}. 
Acoustic mapping is therefore complementary to satellite mapping, extending it into the depth and turbidity range and the fine spatial scale where optical systems stop. More recent acoustic mapping of \textit{Posidonia oceanica} continues in this direction, for example the multi-platform survey of Piazzolla \textit{et al.}, which reconstructed bottom types and seagrass coverage from echosounder data \cite{piazzolla2024integrated}, and Hamouda \textit{et al.} classified benthic habitats, including seagrass beds, from side-scan sonar and sub-bottom profiling in the Egyptian Mediterranean \cite{hamouda2024acoustic}. These acoustic studies, however, rely on manual interpretation or on classical object-based and echosounder classification, and they are often paired with optical data or other sensors.


On the other hand, broader work on seafloor semantic segmentation and seabed sediment classification in general has shifted to deep learning. Burguera and Bonin-Font trained a fully convolutional network for on-line multi-class segmentation of the seafloor into terrain types from an autonomous underwater vehicle, and released a labelled dataset \cite{burguera2020line}. Yang \textit{et al.} designed a multi-channel convolutional network with large kernels to segment SSS images for autonomous navigation \cite{yang2022side}. Zhao \textit{et al.} proposed a dimension-invariant residual network for seabed sediment classification that preserves feature resolution across channels \cite{zhao2023seabed}. Rajani \textit{et al.} introduced a convolutional Vision Transformer for seafloor semantic segmentation and reported state-of-the-art accuracy under real-time constraints \cite{rajani2023}.

A separate and larger body of recent work addresses target and instance segmentation in SSS rather than semantic segmentation. These studies focus on discrete man-made objects such as mines, submarine pipelines, shipwrecks and other structures instead of benthic classes. Huang \textit{et al.} and Tang \textit{et al.} segmented sonar targets with attention-based SOLO and U-Net variants \cite{huang2022dsa, tang2023side}, Wang \textit{et al.} proposed a recurrent pyramid frequency network for instance segmentation of seabed objects \cite{wang2023rpfnet}, and Sethuraman \textit{et al.} released a benchmark for shipwreck segmentation \cite{sethuraman2025machine}. In all of these the aim is to locate sparse targets rather than label continuous seabed cover, distinguishing them from benthic habitat mapping.

A limitation shared across both these groups is that they are trained with dense, pixel-level labels, which are slow and expensive to produce for SSS. This is what limits its use for large-area habitat mapping. The method proposed here removes that obstacle by learning to segment benthic habitats from SSS under image-level supervision only, which makes high-resolution, depth-independent seagrass mapping practical at scale and lets acoustic surveys extend basin-scale optical inventories into the areas they cannot see. To our knowledge, our proposed pipeline is the first to apply weakly supervised semantic segmentation to side-scan sonar, and the first to apply weak supervision to benthic habitat mapping in this modality. The closest work is that of Sledge \textit{et al.}, who used image-level labels for circular-scan synthetic-aperture sonar \cite{sledge2024weakly}, a modality that differs from side-scan sonar in acquisition geometry and resolution and that addresses seafloor and object classes rather than habitats.


\section{Methodology}

We build our approach on the iterative self-improved framework, ISIM \cite{bircanoglu2022isim}, which learns a pixel-level segmentation from image-level labels alone. We retain the overall structure of the framework and adapt two compotenents to the SSS setting. The dense CRF that refines the CAM-derived pseudo-labels is retuned for the low contrast, speckle noise, and weak object boundaries of acoustic imagery, which make the initial CAMs particularly fragmented. The segmentation loss is then reconsidered to address the strong class imbalance of the dataset and the partial supervision provided by incomplete pseudo-labels. The remainder of this section presents the framework and its backbone, the dense CRF refinement, and the segmentation loss, while the training procedure and the handling of class imbalance are described in Section~\ref{sec:training}.

\begin{figure*}[t]
    \centering
    \includegraphics[width=\textwidth]{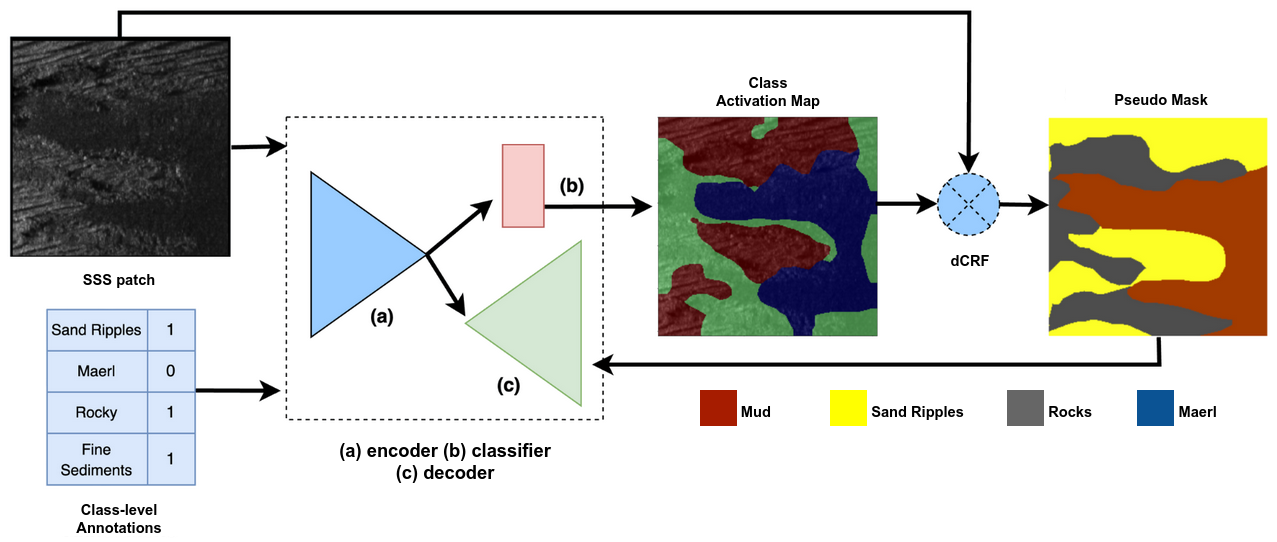}
    \caption{The overall weakly-supervised segmentation framework. (a) denotes the encoder, (b) denotes the classification sub-network, and (c) denotes the decoder.}
    \label{fig:framework}
\end{figure*}

\subsection{Weakly-supervised Segmentation Framework}
\label{sec:wsss}

Figure~\ref{fig:framework} depicts the overall framework. It couples an encoder-decoder segmentation network with a classification sub-network. The classification branch, formed by the encoder and the classification sub-network, is trained for multi-label classification on the image-level labels. Class activation maps (CAMs) \cite{zhou2016learning} are extracted from this branch and thresholded to give initial pseudo-labels, which a dense conditional random field (dCRF) \cite{krahenbuhl2011efficient} then refines to sharpen boundaries and reduce label noise. The refined masks supervise the decoder through a pixel-wise segmentation loss. Training proceeds in an iterative fashion. The classification branch is first trained for a number of epochs, after which CAMs are computed for every training image and converted to pseudo-segmentation labels by the dCRF. These pseudo-labels are then used to train the full encoder-decoder as if they were ground truth. The cycle repeats, so that a stronger segmentation branch yields better CAMs, which in turn yield better pseudo-labels. In this way the framework narrows the supervision gap using only image-level labels and the consistency of the image content. We modify this framework by incorporating our ViT-based encoder-decoder architecture developed for SSS seafloor segmentation \cite{rajani2023}. The classification sub-network sits on top of the encoder features, with a global average pooling (GAP) layer followed by a $1\times1$ convolution that produces the class scores used both for multi-label classification and for CAM extraction.

\subsection{Pseudo-label Refinement}

CAMs indicate object presence but tend to activate only the most discriminative parts of an object, so the pseudo-labels derived from them are often fragmented and poorly aligned with true boundaries. Following the baseline, we use a dense CRF as a post-processing step to turn these CAMs into more coherent masks. The dCRF operates on the raw CAM-derived masks and uses low-level image cues, such as pixel intensity and spatial position, to correct label boundaries by local consistency. Its energy function combines a unary term from the soft CAM outputs, a pairwise bilateral term that encourages nearby pixels of similar appearance and position to share a label, and a spatial smoothness term that penalises isolated label changes. Rather than alter the model, we tune the main dCRF hyperparameters, namely the spatial and colour standard deviations that set the influence radius of the pairwise terms, the weights of the bilateral and spatial kernels, and the number of inference iterations. The effect of this tuning on mask quality is reported in Section~\ref{sec:results}.

\subsection{Segmentation Loss}

In the original framework, the decoder is supervised on the dCRF-refined pseudo-labels with a standard pixel-wise cross-entropy (CE) loss. CE works well under full supervision, but it can underperform under weak supervision, where class imbalance and partial supervision from incomplete pseudo-labels bias training towards the majority class. We therefore study two loss functions that are better suited to this setting. Focal loss \cite{lin2017focal} adds a modulating factor to the cross-entropy that down-weights well-classified pixels and concentrates training on hard examples. Lov\'asz-Softmax loss \cite{berman2018lovasz} optimises the mean intersection-over-union directly through a convex surrogate of the Jaccard index, which aligns the objective with the metric used to evaluate segmentation. We integrate each loss into the segmentation branch on its own, keeping all other training settings fixed, and compare their effect in Section~\ref{sec:results}.

\subsection{Self-supervised Pre-training}
\label{sec:esvit}

Weak supervision reduces but does not remove the need for labelled data, whereas unlabelled SSS imagery is available in far larger quantities. To exploit this, we pre-train the encoder in a self-supervised manner on the unlabelled SSS archive before the weakly-supervised fine-tuning. We adopt EsViT \cite{li2021esvit}, an efficient self-supervised method for vision transformers that builds on the self-distillation framework of DINO \cite{caron2021dino}.

Self-distillation trains a student and a teacher network of identical architecture but different parameters. Several augmented views of an image are generated, comprising global views at high resolution and smaller local views. The student processes all views while the teacher processes only the global ones, and the student is trained to match the teacher output across views, so that the local views are made consistent with the global context. To prevent collapse, the teacher output is centred with a running mean over the mini-batch and sharpened with a temperature, and the teacher parameters are updated as an exponential moving average of the student parameters rather than by gradient descent.

EsViT retains this view-level objective and adds a region-level objective. Besides matching whole views, it matches corresponding local regions between the student and teacher, which encourages the representation to capture fine-grained correspondences between image regions that the view-level task alone tends to lose. This property is desirable for SSS, where discriminative cues are often small and textural. After pre-training, the encoder provides a strong initialisation for both the classification and the segmentation branches of the weakly supervised framework.


\section{Experimental Setup}

\subsection{Dataset}
\label{sec:dataset}

\begin{figure*}[t]
    \centering
    \includegraphics[width=\textwidth]{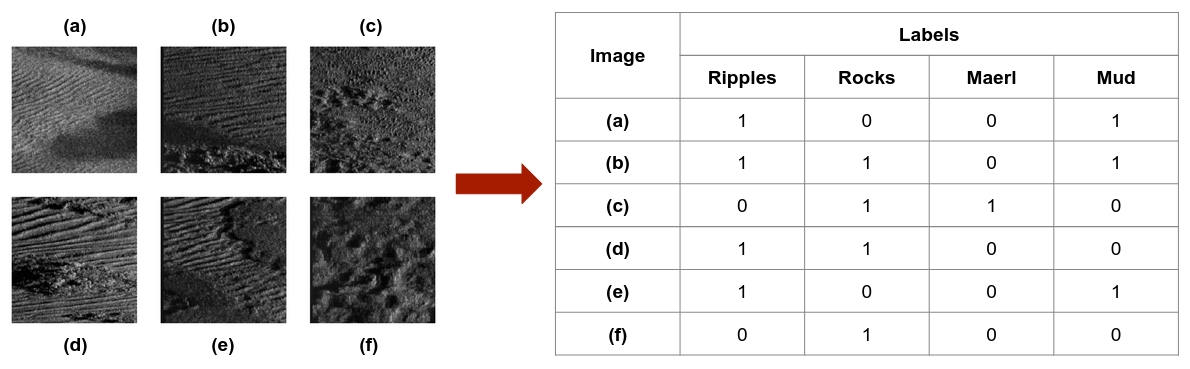}
    \caption{Examples of patches generated from SSS waterfalls and the corresponding image-level ground truth.}
    \label{fig:data}
\end{figure*}

We use the BenthiCat dataset, a large-scale opti-acoustic dataset for benthic classification and habitat mapping~\cite{rajani2026}, to train and evaluate our framework. The dataset contains a subset comprising approximately one million SSS tiles for self-supervised representation learning, collected along the coast of Catalonia, Spain, which covers a wide variety of benthic habitats. This large volume of unannotated SSS data makes it well suited to self-supervised pre-training with DINO. For evaluation, we use the annotated subset of about $36{,}000$ tiles, each provided with a pixel-wise segmentation mask over $12$ benthic habitat classes for supervised fine-tuning. Figure~\ref{fig:data} illustrates some examples of the SSS images along with their multi-class labels for weak supervision.

The raw 12-bit SSS waterfall data underwent several preprocessing steps. First, the data were subjected to logarithmic compression and normalisation to the range $[0,1]$, which stabilised training and preserved low-intensity detail:
\begin{equation}
I'_{\mathrm{normalized}} =
\frac{\ln(1 + I_{\mathrm{raw}})}{\max(\ln(1 + I_{\mathrm{raw}}))}.
\end{equation}

Next, slant-range correction was applied under the flat-floor assumption to convert the slant range $r_s$ to ground range $r_g$:
\begin{equation}
r_g = \sqrt{r_s^2 - h^2},
\end{equation}
where $h$ is the sensor altitude above the seabed; this step removed the nadir compression and the blind zone.

Finally, the data were tiled into overlapping $384 \times 384$ pixel patches with a $192$-pixel stride in both the along- and across-track directions.

\subsection{Training}
\label{sec:training}

The training set is strongly imbalanced. Of about 38{,}000 SSS images, the large majority contain a single class, while roughly 9{,}250 contain two classes and fewer than 550 contain three or more. This imbalance is harmful here because of two connected reasons. First, a CAM is a class-specific weighting of the encoder features, so a sharp CAM requires the classifier to have learned features that separate that class from the others in space. When almost every image is filled by a single class, the multi-label objective can be minimised from global image statistics alone, without ever having to localise one class against a competing one, so the encoder is driven towards image-level recognition rather than spatial class separation. The CAMs it produces are therefore diffuse, and in the few multi-class images the per-pixel argmax across class CAMs collapses onto the dominant class. Second, the minority classes receive few positive gradients, so their class weight vectors stay under-trained and their CAMs remain weak and low in confidence. A fixed threshold then yields sparse or empty pseudo-masks for these classes, or masks that are absorbed by the majority class. Because the encoder is shared with the segmentation branch, these biased pseudo-labels are fed back into training and the bias is amplified across self-training iterations. To counter this, we adopt a random sampling strategy. At each epoch we draw single-class images in proportion to the number of multi-class images in the batch, so that the multi-class examples, which are the only ones that force the network to separate co-occurring classes, contribute a comparable share of the gradient rather than being swamped by the single-class majority.

Alongside the sampling strategy, the dense CRF is tuned as a separate step. Because the dCRF acts only as a post-processing stage on the CAM outputs, its hyperparameters are searched independently of the network weights. We vary the spatial and intensity standard deviations of the pairwise terms, the weights of the bilateral and spatial kernels, and the number of inference iterations, and retain the setting that produces the cleanest refined masks. This configuration is then held fixed while the pseudo-labels are generated for the self-training loop.

All our models were trained on an NVIDIA A100 Tensor Core GPU for 50 epochs with a batch size of 64. We utilized the AdamW optimizer with a weight decay of $1e^{-2}$ and learning rate of $6e^{-5}$, decayed using a polynomial learning rate scheduler with a warm-up of 3 epochs. The CAMs generated by the classification branch are updated every 20 epochs before being refined via dCRF. We also adopted standard data augmentation techniques such as random rotation, random resized crop, random horizontal and vertical flip, random variations in contrast and/or sharpening and Gaussian blur. The models were implemented in PyTorch 1.11.0 and Python 3.8.10. The source code with all hyper-parameter configurations are available at ~\url{https://github.com/CIRS-Girona/w-s3Tseg}.

The self-supervised pre-training, on the other hand, was run for 300 epochs with a learning rate of $5e^{-4}$, decayed using a polynomial learning rate scheduler with a warm-up of 10 epochs. All other hyper-parameter configurations and the source code is available at ~\url{https://github.com/DeeperSense/deepersense-seafloorscan/tree/main/self_supervised/esvit}.

\subsection{Evaluation}
\label{sec:evaluation}

The trained models were initially evaluated on the manually annotated test set. We further carried out field trials using the Girona1000 AUV retrofitted with a Klein 3000 SSS. Small transects were recorded along the port of St. Feliu de Guixols, Spain. This dataset served to demonstrate the generalizability of the model not only to previously unseen data but also to a SSS with diverse configurations and characteristics.


All trained models were then evaluated on an NVIDIA Jetson AGX Orin Developer Kit running Jetpack 5.1.1 with Python 3.8.10 and PyTorch 2.0.0+nv23.5. We report model performance in terms of mean Intersection over Union (mIoU) and inference speed in number of images processed per second (FPS).


\section{Results and Discussion}
\label{sec:results}

\begin{figure*}[t]
    \centering
    \includegraphics[width=\textwidth]{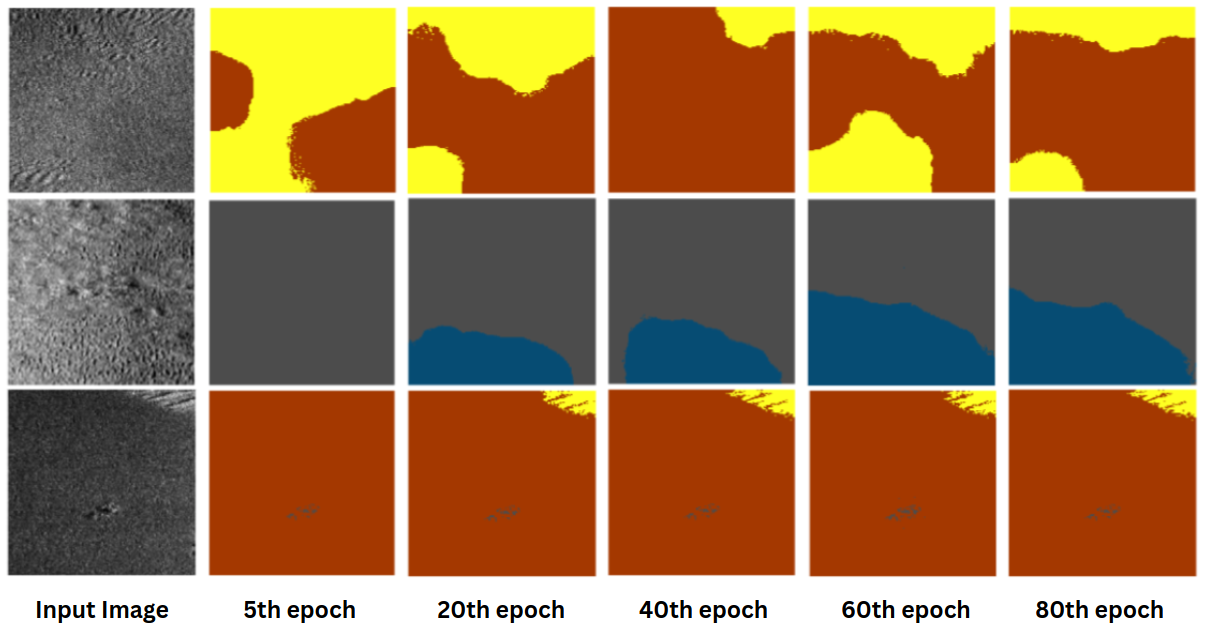}
    \caption{Evolution of pseudo-masks during the course of training.}
    \label{fig:pseudo_masks_progression}
\end{figure*}

Figure~\ref{fig:pseudo_masks_progression} presents the evolution of pseudo-masks during the course of training. As the results depict, the model can identify even the pixels corresponding to small scale objects while also quite accurately capturing inter-class boundaries. This clearly showcases the significant potential of this approach in generating pixel-level annotations from image-level labels. Figures \ref{fig:results1} and \ref{fig:results2} illustrate some of these pixel-level predictions generated by the segmentation branch when iteratively trained using the pseudo masks produced by the classification branch.

The mean IOU between the pixel-level predictions and the pseudo segmentation masks on the test transect was 92.94\%. This suggests that even with an approach that iteratively refines the targets of the segmentation branch over the course of training, the model can generalize well. To further analyse the results, pixel-level ground truth for this test set was manually created. The mean IOU of the pseudo segmentation masks with the ground truth was about 89.3\% whereas that between the pixel-level predictions from the segmentation branch and the ground truth was about 87.6\%. However, self-supervised pre-training resulted in a further increase of about 3\% in mIOU on top of these results.

\begin{figure}[t]
    \centering
    \includegraphics[width=0.4\textwidth]{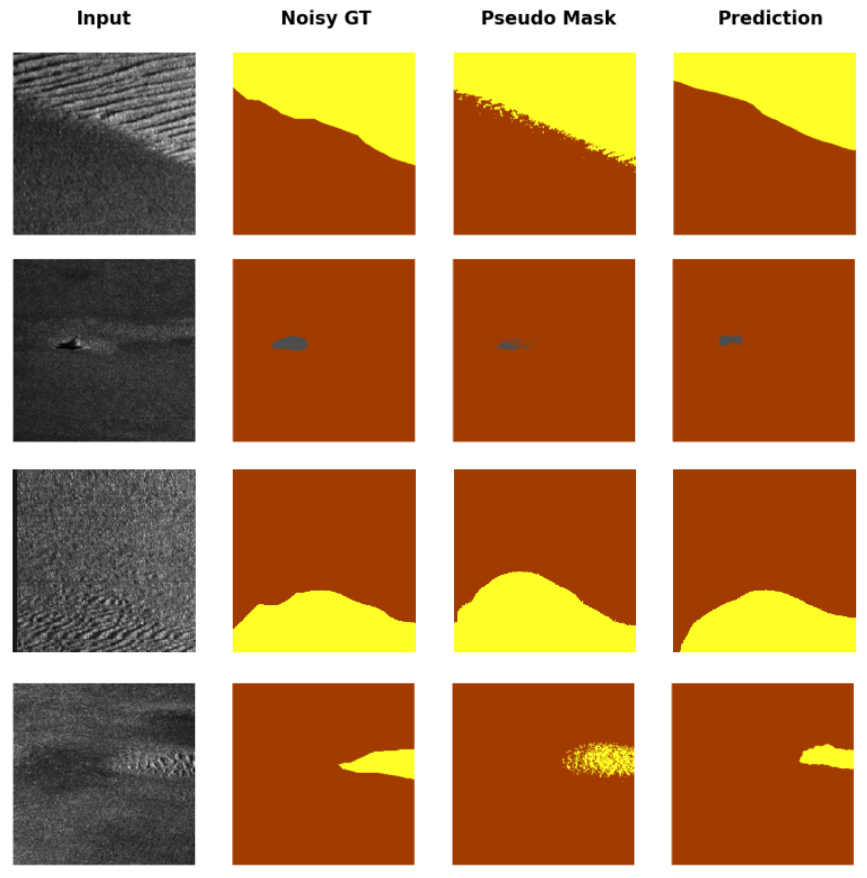}
    \caption{Visualization of predictions on SSS images from the test set.}
    \label{fig:results1}
\end{figure}

\begin{figure}[t]
    \centering
    \includegraphics[width=0.4\textwidth]{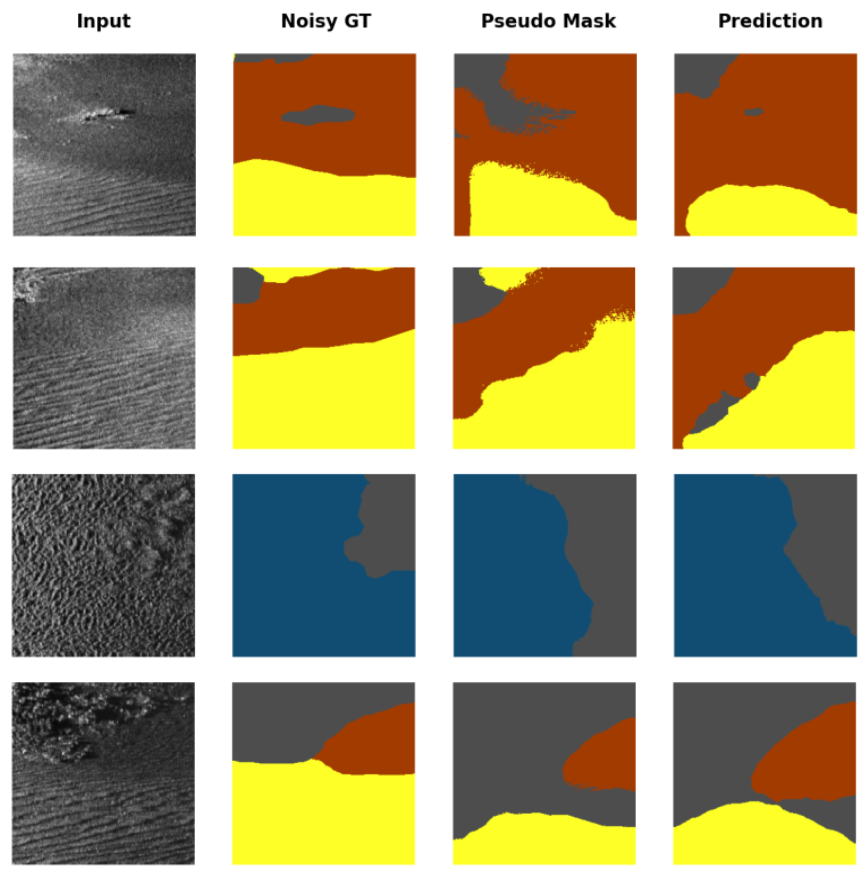}
    \caption{(Continued) Visualization of predictions on SSS images from the test set.}
    \label{fig:results2}
\end{figure}

\begin{figure*}[t]
    \centering
    \includegraphics[width=\textwidth]{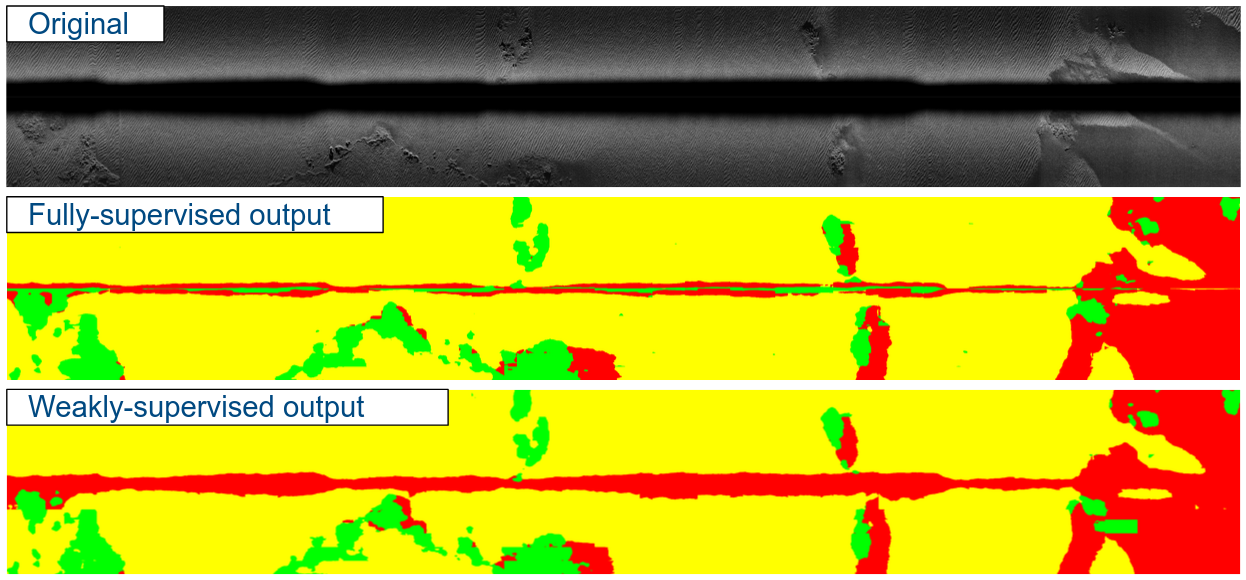}
    \caption{Comparison of results between fully- and weakly-supervised models during field trials.}
    \label{fig:results_field}
\end{figure*}

Figure~\ref{fig:results_field}, on the other hand, depict results from the field tests. Here, a notable discrepancy in classification can be observed along the middle portion of the transect where the port and starboard side images are merged. This discrepancy results from the blind zone and shadows cast near the first bottom return. As a consequence, the fully-supervised model misclassifies these featureless areas as mud and the darker shadowy areas as rocks. The weakly-supervised model, on the other hand, classifies the entire region as mud since it was not explicitly trained to classify shadows as rocks and it therefore associates such dark featureless regions as mud, considering it the closest match.

\section{Conclusion}

We presented a weakly supervised approach to benthic habitat mapping, aimed at reducing the annotation cost that has constrained the use of SSS for large-area seagrass monitoring. By adapting an iterative self-improved framework, the method learns pixel-level segmentation from image-level labels alone, using class activation maps that are refined into pseudo-labels by a dense conditional random field retuned for acoustic imagery. A ViT-based encoder-decoder serves as the backbone, a random sampling strategy mitigates the strong class imbalance of the data, and Lov\'asz-Softmax loss directly optimizes the intersection-over-union resulting in better separation of class boundaries.

On a held-out transect, the segmentation branch reached a mean intersection-over-union of 87.6\% against the ground truth without any pixel-level supervision, close to the 89.3\% of the refined pseudo-labels, and it recovered small-scale objects and inter-class boundaries well. Self-supervised pretraining on unlabelled SSS gave a further improvement of about 3\%. These results indicate that side-scan sonar can be segmented into benthic habitats accurately and at low labelling cost, which makes it a practical complement to optical mapping in the deep and turbid areas that satellites cannot reach.

\section*{Acknowledgments}

This work was supported by Spanish Government through the projects "Intelligent Underwater Robot for Blue Carbon Inventorying (IURBI)" under grant CNS2023-144688 and "Automated Seabed Analysis through Self-Supervised Deep Learning Sonar Technology (ASSiST)" under grant PID2023-149413OB-I00.

\bibliographystyle{IEEEtran}
\bibliography{bibliography}

@article{rajani2026,
  title={Benthicat: An opti-acoustic dataset for advancing benthic classification and habitat mapping},
  author={Rajani, Hayat and Franchi, Valerio and Martinez-Clavel Valles, Borja and Ramos, Raimon and Garcia, Rafael and Gracias, Nuno},
  journal={Earth System Science Data Discussions},
  volume={2026},
  pages={1--32},
  year={2026},
  publisher={G{\"o}ttingen, Germany}
}

@article{rajani2023,
  title={A convolutional vision transformer for semantic segmentation of side-scan sonar data},
  author={Rajani, Hayat and Gracias, Nuno and Garcia, Rafael},
  journal={Ocean Engineering},
  volume={286},
  pages={115647},
  year={2023},
  publisher={Elsevier}
}

@article{fourqurean2012seagrass,
  title={Seagrass ecosystems as a globally significant carbon stock},
  author={Fourqurean, James W and Duarte, Carlos M and Kennedy, Hilary and Marb{\`a}, N{\'u}ria and Holmer, Marianne and Mateo, Miguel Angel and Apostolaki, Eugenia T and Kendrick, Gary A and Krause-Jensen, Dorte and McGlathery, Karen J and others},
  journal={Nature geoscience},
  volume={5},
  number={7},
  pages={505--509},
  year={2012},
  publisher={Nature Publishing Group UK London}
}

@article{waycott2009accelerating,
  title={Accelerating loss of seagrasses across the globe threatens coastal ecosystems},
  author={Waycott, Michelle and Duarte, Carlos M and Carruthers, Tim JB and Orth, Robert J and Dennison, William C and Olyarnik, Suzanne and Calladine, Ainsley and Fourqurean, James W and Heck Jr, Kenneth L and Hughes, A Randall and others},
  journal={Proceedings of the national academy of sciences},
  volume={106},
  number={30},
  pages={12377--12381},
  year={2009},
  publisher={National Academy of Sciences}
}

@article{pasqualini1998mapping,
  title={Mapping ofposidonia oceanicausing aerial photographs and side scan sonar: Application off the island of corsica (france)},
  author={Pasqualini, V and Pergent-Martini, C and Clabaut, P and Pergent, G},
  journal={Estuarine, Coastal and Shelf Science},
  volume={47},
  number={3},
  pages={359--367},
  year={1998},
  publisher={Elsevier}
}

@book{blondel2009handbook,
author = {Blondel, Philippe},
title = {The Handbook of Sidescan Sonar},
publisher = {Springer},
address = {Berlin, Heidelberg},
year = {2009}
}

@inproceedings{kirillov2023segment,
  title={Segment anything},
  author={Kirillov, Alexander and Mintun, Eric and Ravi, Nikhila and Mao, Hanzi and Rolland, Chloe and Gustafson, Laura and Xiao, Tete and Whitehead, Spencer and Berg, Alexander C and Lo, Wan-Yen and others},
  booktitle={2023 IEEE/CVF international conference on computer vision (ICCV)},
  pages={3992--4003},
  year={2023},
  organization={IEEE}
}

@article{katija2022fathomnet,
  title={FathomNet: A global image database for enabling artificial intelligence in the ocean: K. Katija et al.},
  author={Katija, Kakani and Orenstein, Eric and Schlining, Brian and Lundsten, Lonny and Barnard, Kevin and Sainz, Giovanna and Boulais, Oceane and Cromwell, Megan and Butler, Erin and Woodward, Benjamin and others},
  journal={Scientific reports},
  volume={12},
  number={1},
  pages={15914},
  year={2022},
  publisher={Nature Publishing Group UK London}
}

@article{chen2025weakly,
  title={Weakly-supervised semantic segmentation with image-level labels: from traditional models to foundation models},
  author={Chen, Zhaozheng and Sun, Qianru},
  journal={ACM Computing Surveys},
  volume={57},
  number={5},
  pages={1--29},
  year={2025},
  publisher={ACM New York, NY}
}

@inproceedings{ahn2018learning,
  title={Learning pixel-level semantic affinity with image-level supervision for weakly supervised semantic segmentation},
  author={Ahn, Jiwoon and Kwak, Suha},
  booktitle={2018 IEEE/CVF Conference on Computer Vision and Pattern Recognition},
  pages={4981--4990},
  year={2018},
  organization={IEEE}
}

@article{krahenbuhl2011efficient,
  title={Efficient inference in fully connected crfs with gaussian edge potentials},
  author={Kr{\"a}henb{\"u}hl, Philipp and Koltun, Vladlen},
  journal={Advances in neural information processing systems},
  volume={24},
  year={2011}
}

@inproceedings{lin2017focal,
  title={Focal loss for dense object detection},
  author={Lin, Tsung-Yi and Goyal, Priya and Girshick, Ross and He, Kaiming and Doll{\'a}r, Piotr},
  booktitle={Proceedings of the IEEE international conference on computer vision},
  pages={2980--2988},
  year={2017}
}

@inproceedings{berman2018lovasz,
  title={The Lov{\'a}sz-Softmax loss: A tractable surrogate for the optimization of the intersection-over-union measure in neural networks},
  author={Berman, Maxim and Rannen Triki, Amal and Blaschko, Matthew B},
  booktitle={Proceedings of the IEEE Conference on Computer Vision and Pattern Recognition},
  pages={4413--4421},
  year={2018}
}

@article{traganos2022spatially,
  title={Spatially explicit seagrass extent mapping across the entire mediterranean},
  author={Traganos, Dimosthenis and Lee, Chengfa Benjamin and Blume, Alina and Poursanidis, Dimitris and {\v{C}}i{\v{z}}mek, Hrvoje and Deter, Julie and Ma{\v{c}}i{\'c}, Vesna and Montefalcone, Monica and Pergent, G{\'e}rard and Pergent-Martini, Christine and others},
  journal={Frontiers in Marine Science},
  volume={9},
  pages={871799},
  year={2022},
  publisher={Frontiers Media SA}
}

@article{gimenez2025generalizable,
  title={A generalizable deep learning framework for large-scale mapping of seagrass habitats},
  author={Gim{\'e}nez-Romero, {\`A}lex and Ferchichi, Dhafer and Moreno-Spiegelberg, Pablo and Sintes, Tom{\`a}s and Mat{\'\i}as, Manuel A},
  journal={Ecological Indicators},
  volume={180},
  pages={114349},
  year={2025},
  publisher={Elsevier}
}

@article{jeon2021semantic,
  title={Semantic segmentation of seagrass habitat from drone imagery based on deep learning: A comparative study},
  author={Jeon, Eui-ik and Kim, Sunghak and Park, Soyoung and Kwak, Juwon and Choi, Imho},
  journal={Ecological Informatics},
  volume={66},
  pages={101430},
  year={2021},
  publisher={Elsevier}
}

@article{pasqualini2000contribution,
  title={Contribution of side scan sonar to the management of Mediterranean littoral ecosystems},
  author={Pasqualini, V and Clabaut, P and Pergent, G and Benyoussef, L and Pergent-Martini, C},
  journal={International Journal of Remote Sensing},
  volume={21},
  number={2},
  pages={367--378},
  year={2000},
  publisher={Taylor \& Francis}
}

@article{piazzolla2024integrated,
  title={An integrated approach for the benthic habitat mapping based on innovative surveying technologies and ecosystem functioning measurements},
  author={Piazzolla, Daniele and Scanu, Sergio and Mancuso, Francesco Paolo and Bosch-Belmar, Mar and Bonamano, Simone and Madonia, Alice and Scagnoli, Elena and Tantillo, Mario Francesco and Russi, Martina and Savini, Alessandra and others},
  journal={Scientific Reports},
  volume={14},
  number={1},
  pages={5888},
  year={2024},
  publisher={Nature Publishing Group UK London}
}

@article{hamouda2024acoustic,
  title={Acoustic-based classification of marine geophysical data for benthic habitat mapping in the littoral zone of Qaitbay Citadel of Alexandria},
  author={Hamouda, Amr Z and Fekry, Ahmed and El-Gharabawy, Suzan},
  journal={Egyptian Journal of Aquatic Research},
  volume={50},
  number={1},
  pages={8--16},
  year={2024},
  publisher={Elsevier}
}

@article{burguera2020line,
  title={On-line multi-class segmentation of side-scan sonar imagery using an autonomous underwater vehicle},
  author={Burguera, Antoni and Bonin-Font, Francisco},
  journal={Journal of Marine Science and Engineering},
  volume={8},
  number={8},
  pages={557},
  year={2020},
  publisher={MDPI}
}

@article{yang2022side,
  title={Side-scan sonar image segmentation based on multi-channel CNN for AUV navigation},
  author={Yang, Dianyu and Cheng, Chensheng and Wang, Can and Pan, Guang and Zhang, Feihu},
  journal={Frontiers in Neurorobotics},
  volume={16},
  pages={928206},
  year={2022},
  publisher={Frontiers Media SA}
}

@article{zhao2023seabed,
  title={Seabed sediments classification based on side-scan sonar images using dimension-invariant residual network},
  author={Zhao, Yuxin and Zhu, Kexin and Zhao, Ting and Zheng, Liangfeng and Deng, Xiong},
  journal={Applied Ocean Research},
  volume={130},
  pages={103429},
  year={2023},
  publisher={Elsevier}
}

@article{huang2022dsa,
  title={DSA-SOLO: double split attention SOLO for side-scan sonar target segmentation},
  author={Huang, Honghe and Zuo, Zhen and Sun, Bei and Wu, Peng and Zhang, Jiaju},
  journal={Applied Sciences},
  volume={12},
  number={18},
  pages={9365},
  year={2022},
  publisher={MDPI}
}

@article{tang2023side,
  title={Side-scan sonar underwater target segmentation using the BHP-UNet},
  author={Tang, Yulin and Wang, Liming and Li, Houpu and Bian, Shaofeng},
  journal={EURASIP Journal on Advances in Signal Processing},
  volume={2023},
  number={1},
  pages={76},
  year={2023},
  publisher={Springer}
}

@article{wang2023rpfnet,
  title={RPFNet: Recurrent pyramid frequency feature fusion network for instance segmentation in side-scan sonar images},
  author={Wang, Zhen and Zhang, Shanwen and Zhang, Chuanlei and Wang, Buhong},
  journal={IEEE Journal of Selected Topics in Applied Earth Observations and Remote Sensing},
  year={2023},
  publisher={IEEE}
}

@article{sethuraman2025machine,
  title={Machine learning for shipwreck segmentation from side scan sonar imagery: Dataset and benchmark},
  author={Sethuraman, Advaith V and Sheppard, Anja and Bagoren, Onur and Pinnow, Christopher and Anderson, Jamey and Havens, Timothy C and Skinner, Katherine A},
  journal={The International Journal of Robotics Research},
  volume={44},
  number={3},
  pages={341--354},
  year={2025},
  publisher={SAGE Publications Sage UK: London, England}
}

@article{sledge2024weakly,
  title={Weakly-Supervised Semantic Segmentation of Circular-Scan, Synthetic-Aperture-Sonar Imagery},
  author={Sledge, Isaac J and Byrne, Dominic M and King, Jonathan L and Ostertag, Steven H and Woods, Denton L and Prater, James L and Kennedy, Jermaine L and Marston, Timothy M and Principe, Jose C},
  journal={arXiv preprint arXiv:2401.11313},
  year={2024}
}

@article{bircanoglu2022isim,
  title={Isim: Iterative self-improved model for weakly supervised segmentation},
  author={Bircanoglu, Cenk and Arica, Nafiz},
  journal={arXiv preprint arXiv:2211.12455},
  year={2022}
}

@article{li2021esvit,
  title={Efficient self-supervised vision transformers for representation learning},
  author={Li, Chunyuan and Yang, Jianwei and Zhang, Pengchuan and Gao, Mei and Xiao, Bin and Dai, Xiyang and Yuan, Lu and Gao, Jianfeng},
  journal={arXiv preprint arXiv:2106.09785},
  year={2021}
}

@inproceedings{caron2021dino,
  title={Emerging properties in self-supervised vision transformers},
  author={Caron, Mathilde and Touvron, Hugo and Misra, Ishan and J{\'e}gou, Herv{\'e} and Mairal, Julien and Bojanowski, Piotr and Joulin, Armand},
  booktitle={2021 IEEE/CVF international conference on computer vision (ICCV)},
  pages={9630--9640},
  year={2021},
  organization={IEEE}
}

@inproceedings{zhou2016learning,
  title={Learning deep features for discriminative localization},
  author={Zhou, Bolei and Khosla, Aditya and Lapedriza, Agata and Oliva, Aude and Torralba, Antonio},
  booktitle={Proceedings of the IEEE conference on computer vision and pattern recognition},
  pages={2921--2929},
  year={2016}
}

\end{document}